\documentclass[letterpaper, 10 pt, conference]{ieeeconf}  
\pdfoutput=1
\IEEEoverridecommandlockouts                              % This command is only needed if you want to use the \thanks command
\usepackage{graphicx} % for pdf, bitmapped graphics files
\usepackage{amsmath} % assumes amsmath package installed
\usepackage{amssymb}  % assumes amsmath package installed
\usepackage[T1]{fontenc}
\usepackage{url}

\usepackage{color} 
\usepackage{xcolor}

\newcommand{\vale}[1]{\textcolor{black}{#1}}

\newcommand{\revised}[1]{\textcolor{black}{#1}}
\newcommand{\finalrev}[1]{\textcolor{black}{#1}}

\usepackage{acronym}
\acrodef{FIERL}{Fault Identification Enhancement with Reinforcement Learning}

\acrodef{FDI}{Fault Detection and Identification}
\acrodef{PFD}{Passive Fault Detection}
\acrodef{AFD}{Active Fault Detection}
\acrodef{AFDC}{Active Fault Detection and Control}

\acrodef{RL}{Reinforcement Learning}
\acrodef{MDP}{Markov Decision Process}
\acrodef{CRL}{Constrained Reinforcement Learning}
\acrodef{CMDP}{Constrained Markov Decision Process}
\acrodef{CPO}{Constrained Policy Optimization}

\acrodef{MLP}{Multi-Layer Perceptron}
\acrodef{GAE}{Generalized Advantage Estimation}

\title{\LARGE \bf \acf{FIERL}}

\author{Valentina Zaccaria*, Davide Sartor*, Simone Del Favero and Gian Antonio Susto% <-this % stops a space
\thanks{V. Zaccaria and D. Sartor equally contributed to this work. All authors are with the Department of Information Engineering, University of Padua, Padua PD 35131, Italy. This study was partially carried out within the MICS (Made in Italy – Circular and Sustainable) Extended Partnership and received funding from Next-GenerationEU (Italian PNRR – M4 C2, Invest 1.3 – D.D. 1551.11-10-
2022, PE00000004).}}

\begin{document}

\maketitle
\thispagestyle{empty}
\pagestyle{empty}

\begin{abstract}
This letter presents a novel approach in the field of \ac{AFD}, by explicitly separating the task into two parts: \ac{PFD} and control input design. 
This formulation is very general, and most existing \ac{AFD} literature can be viewed through this lens. By recognizing this separation, \ac{PFD} methods can be leveraged to provide components that make efficient use of the available information, while the control input is designed in order to optimize the gathering of information. 
The core contribution of this work is \ac{FIERL}, a general simulation-based approach for the design of such control strategies, using \ac{CRL} to optimize the performance of arbitrary passive detectors. The control policy is learned without the need of knowing the passive detector inner workings, making \ac{FIERL} broadly applicable. However, it is especially useful when paired with the design of an efficient passive component. 
Unlike most \ac{AFD} approaches, \ac{FIERL} can handle fairly complex scenarios such as continuous sets of fault modes. The effectiveness of \ac{FIERL} is tested on a benchmark problem for actuator fault diagnosis, where \ac{FIERL} is shown to be fairly robust, being able to generalize to fault dynamics not seen in training.
\end{abstract}

\section{Introduction}
\label{sec:introduction}
In automated systems, faults can arise for a wide variety of reasons, resulting in disruptions of the smooth operation of the process and in safety hazards. 
In this scenario, the need for reliable \ac{FDI} modules has arisen. Indeed, detecting faults at the earliest stage enables the implementation of an appropriate recovery strategy, preventing the propagation of faults and the potential subsequent failure of the system, thereby avoiding severe safety, economic or environmental consequences.

\ac{FDI} methods can be broadly categorized into two main groups: \acf{PFD} and \acf{AFD}, depending on whether or not interactions with the system are possible.
In \ac{PFD} methods, process data are collected and analyzed without interfering with the system and anomalous system behaviour is identified by relying on nominal input-output relations, \revised{which can be either derived from a model or learned from data.} 
Model-based \ac{PFD} is extensively explored in the literature \cite{ding2013model}, including a variety of well-established approaches, most notably residual- or observer-based strategies \cite{PATTON1997671}.
Most recently, data-driven methods have also been proposed,
achieving successful outcomes in various fields \cite{MdNorCheHassanHussain+2020+513+553, s21124024}.
\\
\indent \revised{However, in some scenarios \ac{PFD} may yield suboptimal diagnosis performances. For instance, this can occur in the presence of incipient faults, that may be masked by noise in the real system, or due to the effect of closed-loop feedback controllers that tend to compensate, to some degree, for system deviations caused by faults.}

\revised{These issues can be mitigated with the use of \ac{AFD} approaches. Such methodologies involve the design of a minimally intrusive input, called \textit{auxiliary input}, that is added to the primary control action. This can improve the diagnosis by gathering more informative data, but may result in a degradation in the control performance \cite{forouzanfar2017constrained}.
This trade-off is better handled by integrated \ac{AFDC} techniques that jointly design the primary control and the auxiliary input, at the cost of significantly increased complexity.}

\revised{Both \ac{AFD} and \ac{AFDC} techniques commonly assume the set of fault modes to be finite, in order to make the problem tractable. In this case, the goal is to achieve separation of the different fault dynamics. 
This can be theoretically guaranteed by assuming deterministic and bounded disturbances \cite{MARSEGLIA2017223, XU2021109558, TAN2021109602}, but it is possible to model more realistic processes, by considering stochastic disturbances and moving to probabilistic objectives, like minimizing the Bayesian risk of model misdiagnosis \cite{paulson2017input, 6819827}, or maximizing a statistical distance measure of the predicted output distributions of different fault modes \cite{8431031,HEIRUNG201715934, https://doi.org/10.1002/rnc.3627, TAN2022110348}.}
\revised{The aforementioned formulations usually involve nested optimization problems, whose solution via traditional methods require significant simplifications or heavily restricting conditions, especially in the case of \ac{AFDC}. 
For this reason, recent studies have started to investigate the use of \ac{RL} for solving these optimization problems, with promising results \cite{9693676,https://doi.org/10.1002/acs.3456,YAN2023364, 7798581}. }

\revised{In this work, we introduce \ac{FIERL}, a general framework for \ac{RL}-based \ac{AFD}. In \ac{FIERL}, the \ac{AFD} (and \ac{AFDC}) paradigm is reformulated by explicitly disentangling two key tasks: the first is a \ac{PFD} step, which handles the diagnosis itself, the second is the input design, that aims to maximize the performance of the \ac{PFD} component.
While most \ac{AFD} techniques in the literature could be viewed through this lens, when using \ac{RL} this separation becomes particularly useful to recognize.
In this case the \ac{PFD} component can be treated as a generic black box, allowing for seamless integration of any \ac{PFD} strategy into the optimization loop. This not only makes \ac{FIERL} broadly applicable, but also allows it to tackle complex diagnostic scenarios (i.e. continuous spectrum of faults).
Moreover, \ac{FIERL} can naturally be applied to integrated \ac{AFDC}, which is therefore the main focus of this work. However, the discussion could be easily adapted to \ac{AFD} without control.}

This letter unfolds as follows: Section \ref{sec:preliminaries} covers the preliminary notions required, Section \ref{sec:contrained reinforcement learning} briefly describes the basics of constrained \ac{RL} up to the motivation and derivation of the policy optimization algorithm considered in this work. Section \ref{sec:framework general} describes in detail a general formulation of the proposed strategy, while section \ref{sec:framework application} discusses an example application considering the class of linear systems affected by actuator faults. Finally, section \ref{sec:case study} provides some experimental results, evaluating the performance improvements of using \ac{RL} over a naive approach, followed by brief closing remarks in \ref{sec:conclusions}, \revised{discussing in detail the advantages of the proposed approach as well as its limitations}.

\begin{figure*}
    \center
    \includegraphics[width=0.82\textwidth]{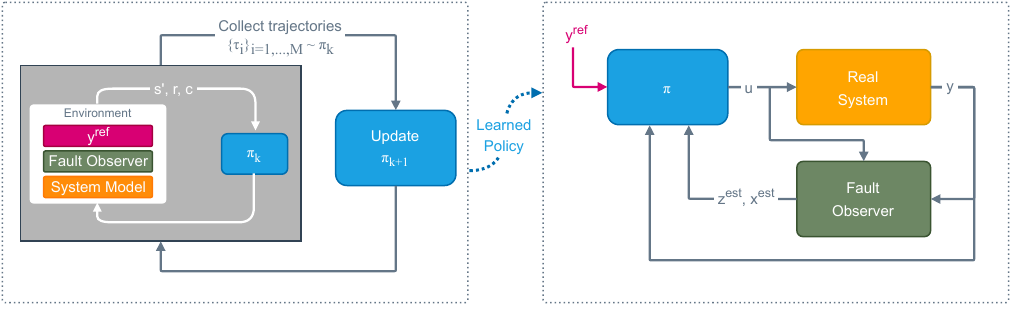}
    \caption{\revised{\ac{FIERL}: depiction of the offline policy training phase (left) and of the control flow during deployment (right).}}
    \label{fig:framework}
\end{figure*}

\section{Preliminaries}
\label{sec:preliminaries}
\subsection{System Modelling and Notation}
\label{subsec:system modelling}

Consider generic time-varying faults $z_t\in Z$ that affect the dynamics of a discrete-time stochastic system: 
\begin{equation}
\label{eq:system equations general}
\begin{aligned}
x_{t+1} &= g(z_t, x_t, u_t, w_t), \\
y_t &= h(z_t, x_t, u_t, v_t), \\
z_{t+1} &= l(z_t,x_t, u_t),
\end{aligned}
\end{equation}
where ${x_t\in\mathbb{R}^{n_x}}$, ${y_t\in\mathbb{R}^{n_y}}$ and ${u_t\in\mathbb{R}^{n_u}}$ represent respectively the system state, output and input at time $t$, ${v_t\in\mathbb{R}^{n_v} \sim \mathcal{N}(\mu_{v}, \Sigma_{v})}$ and ${w_t\in\mathbb{R}^{n_w} \sim \mathcal{N}(\mu_{w}, \Sigma_{w})}$ represent independent \revised{stationary} Gaussian measurement and process noises, the functions $g(\cdot)$ and $h(\cdot)$ describe, respectively, the system dynamics and model output and notably depend on the fault $z_t$. The fault free mode is given by a specific element ${z^0\in Z}$ and the fault evolution dynamics are described by the stochastic function $l(\cdot)$.

\finalrev{In our example application, we consider linear systems with actuator faults, where Eq. \ref{eq:system equations general} is reduced to:}
\begin{equation}
    \label{eq:system equations linear}
    \begin{aligned} 
    x_{t+1} &= A x_{t} + B \operatorname{diag}(z_{t}) u_{t} + w_t \\
    y_{t} &= C x_{t} + v_t\\
    z_{t+1} &= z_t + \xi_t
    \end{aligned}
\end{equation}
where $A$, $B$ and $C$ are known real-valued matrices and each component of $z_t\in\mathbb{R}^{n_u}$ models partial faults of the corresponding actuator. Fault evolution is independent of $x_t$ and $u_t$, and is characterized by the random variable $\xi_t$.

\subsection{Reinforcement Learning}
\label{subsec:reinforcement learning}
In basic \acf{RL} formulation, an \textit{agent} interacts with its \textit{environment}, trying to maximize some notion of \textit{cumulative reward}.
Basic \finalrev{sequential} decision-making problems of this kind are usually modelled as \ac{MDP}. 
\revised{The structural components of an \ac{MDP} are fully represented by the tuple ${\left\langle \mathcal{S}, \mathcal{A}, \mathcal{P}, \mathcal{R} \right\rangle}$, where the state space $S$ and action space $A$ are arbitrary sets. The environment state ${S_t \in \mathcal{S}}$ follows a discrete-time stochastic process, defined by the state transition function ${\mathcal{P}\left(s, a, s'\right) = \mathbb{P}\left[S_{t+1}=s' \mid S_t=s, A_t=a\right]}$ which depends on the action ${A_t \in \mathcal{A}}$ performed by the agent. Finally, the reward function ${\mathcal{R}(s, a, s')}$ is used to measure the quality of a given state transition.}
The behaviour of the agent is generally characterized by a \textit{stochastic policy}, a function  \finalrev{${\pi:\mathcal{S}\rightarrow \mathbf{P}(\mathcal{A})}$} that specifies the probability that a given action will be selected given the environment state. 
The realization of an action state-trajectory ${\tau=(s_0,a_0, s_{1},a_{1}, ...)}$ obtained by following policy $\pi$ is commonly denoted by ${\tau \sim \pi}$.
The return (cumulative reward) represents the total reward (discounted by ${\gamma \in (0,1]}$) accumulated by the agent over the trajectory $\tau$.
The goal of \ac{RL} is to find the optimal policy amongst \finalrev{a} set of admissible ones $\Pi$: 
\begin{equation}
    \pi^* = \arg \max_{\pi \in \Pi} \mathbb{E}_{\tau \sim \pi} \left[ \sum_{t=0}^\infty \gamma^t \mathcal{R}(s_t, a_t, s_{t+1}) \right].
\end{equation}
Many common algorithms also make use of the value function $V^\mathcal{R}_\pi$, the action-value function $Q^\mathcal{R}_\pi$ and the advantage function $A^\mathcal{R}_\pi$ defined as:
\begin{equation}
\begin{aligned}
&V^\mathcal{R}_\pi(s) = \mathbb{E}_{\tau \sim \pi} \left[ \sum_{t=0}^\infty \gamma^t \mathcal{R}(s_t, a_t, s_{t+1}) \mid s_0 = s\right], \\
&Q^\mathcal{R}_\pi(s,a) = \mathbb{E}_{\tau \sim \pi} \left[\sum_{t=0}^\infty \gamma^t \mathcal{R}(s_t, a_t, s_{t+1}) \mid s_0 = s, a_0 = a\right], \\
&A^\mathcal{R}_\pi(s,a) = Q_\pi(s,a) - V_\pi(s).
\end{aligned}
\end{equation}

\section{Constrained Reinforcement Learning}
\label{sec:contrained reinforcement learning}

\acf{CRL} is the general problem of training an \ac{RL} agent under specific constraints, usually with the purpose of satisfying them throughout both training and testing stages, ensuring that the agent operates within predefined boundaries.
\revised{\ac{CRL} can also be used to handle conflicting objectives, by framing one of the two as a constraint and optimizing for the other.}

The de-facto standard to characterize \ac{CRL} is the \ac{CMDP} \cite{altman-constrainedMDP}. A \ac{CMDP} extends a \ac{MDP} including a set of $m$ cost functions ${C_i: \mathcal{S}\times\mathcal{A}\times\mathcal{S}\rightarrow\mathbb{R}}$.
Costs have the same structure as rewards, therefore one can define the expected return of $C_i$, defining the \finalrev{expected} \textit{constraint return} (or \textit{$C_i$-return}):
\begin{equation}
J_{C_i}(\tau) = \mathbb{E}_{\tau \sim \pi} \left[\sum_{t=0}^\infty \gamma^t C_i(s_t,a_t,s_{t+1})\right].
\end{equation}
The optimal policy in a \ac{CMDP} is obtained from the constrained optimization problem: 
\begin{equation}
    \label{cmdp_optimal_policy}
    \pi^* = \underset{\substack{\pi : J_{C_i}(\pi) \leq d_i, \\ i=1,\ldots,m}}{\arg\max} J_{\mathcal{R}}(\pi),
\end{equation}
where $d_i$ are hyper-parameters representing individual constraints.
This is equivalent to the optimal policy of the simple \ac{MDP}, where the set of feasible policies is restricted to:
\begin{equation}
    \Pi_C = \{\pi : J_{C_i}(\pi)\leq d_i, i=1, ...,m\}.
\end{equation}

\subsection{Constrained Policy Optimization}
\label{subsec:cpo}

\ac{CPO} \cite{achiam2017constrained} is a general-purpose local policy search algorithm for \ac{CRL}. It has theoretical guarantees for near-constraint satisfaction throughout all training and it has demonstrated to be effective in learning neural network policies for complex control tasks with high-dimensional constraints.
One of the main advantages of CPO is that
\revised{it generally outperforms other methods like primal-dual optimization \cite{DBLP:journals/corr/ChowGJP15} on enforcing constraints without significantly compromising performance with respect to returns \cite{achiam2017constrained}.} \ac{CPO} shows to outperform fixed-penalty methods that incorporate constraints directly in the reward function.
\ac{CPO} is a local policy search method and as such\revised{, given a parametrization $\theta$ of the policy space, it aims to find an approximation to the optimal policy} by iterated optimization 
of a surrogate objective $J(\pi)$ within a local region of radius $\delta$ in the policy parameter space defined by 
\revised{some distance measure $D$}:
\begin{equation}
    \begin{aligned}
     \pi_{\theta_{k+1}} = &\arg\max_{\pi\in\Pi} J(\pi).                    \\
     \text{s.t.} \quad & D(\pi,\pi_{\theta_k}) \leq \delta
    \end{aligned}
\end{equation} 
When dealing with \ac{CMDP}, the policy domain $\Pi$ is replaced by the set of feasible policies $\Pi_C$.
Unfortunately, assessing which policies lie in  $\Pi_C$ requires precise off-policy evaluation, making the local update impractical.
To overcome this issue, \ac{CPO} uses a principled approximation of the objective and the constraints that depends on the state distribution under the current policy (denoted $s\sim\pi_{\theta_k}$) and can therefore be estimated using on-policy samples, as follows
\begin{equation}
\label{eq:local policy update CPO}
    \begin{aligned}
     \pi_{\theta_{k+1}} = &\arg\max_{\pi\in\Pi} \quad \mathbb{E}_{\substack{s\sim \pi_{\theta_k} \\ a\sim\pi_{\theta_k}}}\left[A^\mathcal{R}_{\pi_{\theta_k}}(s,a)\right]. \\
                    \text{s.t.} \quad &J_{C_i}(\pi_{\theta_k}) + \dfrac{1}{1-\gamma}  \mathbb{E}_{\substack{s\sim \pi_{\theta_k} \\ a\sim \pi_{\theta_k}}}\left[A^{C_i}_{\pi_{\theta_k}}(s,a)\right]
                    \leq d_i \\
                    & \mathbb{E}_{s\sim\pi_k}\left[D_{KL}(\pi_{\theta}(s) \Vert \pi_{\theta_k}(s))\right] \leq \delta
    \end{aligned}  
\end{equation} 
The use of a constraint based on the Kullback-Leibler-divergence ${D}_{KL}$ \revised{\cite{kullback1951information}} makes \ac{CPO} a \textit{trust region} method, allowing for theoretical guarantees on monotonic performance improvement and approximate satisfaction of constraints. \revised{These aspects are discussed in detail in \cite{achiam2017constrained}}. 
%\forse{This is crucial when dealing with neural network policies, as they have been shown to be subject to performance collapse following bad updates \cite{10.5555/3045390.3045531}.}
Solving directly the \ac{CPO} update in Eq. \ref{eq:local policy update CPO} is intractable when dealing with thousands of parameters. An efficient implementation relies on a linearization of both the objective and the cost constraint around the current policy $\pi_{\theta_k}$ while the KL-divergence is approximated by second-order expansion.

\revised{Due to all these approximations, the resulting iterates might be unfeasible and therefore the implementation of a suitable recovery strategy is necessary.
All the technical implementation details are described in \cite{achiam2017constrained}}.

\section{\acl{FIERL}}
\label{sec:framework general}

In this work the \ac{AFD} paradigm is reinterpreted as the enhancement of \ac{PFD} approaches, by designing a feasible control strategy that optimizes the passive component accuracy. Indeed, \ac{AFD} is composed of two distinct tasks: the selection of a control action, and the computation of the fault detector output signal $z_t^{est}$. Deriving the optimal detector output, based on the available information, is essentially a \ac{PFD} task. This is why most \ac{AFD} (and by extension \ac{AFDC}) techniques, which aim to optimize some metric related to fault diagnosis accuracy, often employ nested optimization loops. 
\revised{For instance, for many set-separation based approaches the inner optimization performs template matching which can be thought of as an implicit \ac{PFD} step.} 

\revised{In \ac{FIERL}, we explicitly disentangles these two tasks, relying on existing approaches for the \ac{PFD} part, and focusing on the input design, using \ac{RL} to find a control strategy that improves the accuracy of the selected \ac{PFD} module.}
\revised{In particular, we propose the use of \ac{CRL} to handle the conflicting objectives of diagnosis accuracy and control quality. Instead of solving a dual-objective optimization problem, we directly optimize for the accuracy of the diagnosis, under some constraints on control performance. The rationale being that fixing the desired control performance bounds via the hyper-parameters $d_i$ is generally more intuitive than balancing the conflicting objectives directly. From these considerations, the \ac{AFD} problem can be formulated as a \ac{CMDP} where:
\begin{itemize}
    \item The \textbf{environment state} should contain \finalrev{$x$, $z$} and the passive component estimates $x^{est}_t$ and $z^{est}_t$ for system state and current fault. In the case of \ac{AFDC}, information related to the control task, like the reference signal $y^{ref}_t$ or system output $y_t$, should also be included. \finalrev{}
    \item The \textbf{agent action} $a_t$ can be either the control input $u_t$ in the case of integrated \ac{AFDC}, or an auxiliary control signal in the case of \ac{AFD} without control;
    \item The \textbf{reward function} is defined in terms of metric of choice $\mathcal{M}$, that expresses the instantaneous adherence of the fault estimate signal $z^{est}_t$ to the true fault $z_t$.    
    ${\mathcal{R}(s_t, a_t, s_{t+1}) = \mathcal{M}(z^{est}_t, z_t)}$
    \item The \textbf{cost function} is used to enforce the desired guarantees on control. In this work we focus on enforcing adherence of the system output $y_t$ to a reference signal $y_t^{ref}$, using a distance metric of choice $\mathcal{D}$. 
    ${\mathcal{C}(s_t, a_t, s_{t+1}) = \mathcal{D}(y^{ref}_t, y_t)}$.
\end{itemize}
}
\noindent\revised{Therefore, the optimal control policy derived in \ref{cmdp_optimal_policy}, becomes:}
\begin{equation}
    \label{eq:optimal policy framework}
    \centering
    \begin{aligned}
    \pi_{\theta^*} = &\arg\max_{\pi_\theta} \mathbb{E}_{\tau\sim\pi_\theta}\left[\sum_{t=0}^{\infty} \gamma^t \mathcal{M}(z^{est}_{t}, z_{t})\right]. \\
    & \text{s.t.} \quad \mathbb{E}_{\tau\sim\pi_\theta}\left[\sum_{t=0}^{\infty} \gamma_c^t \mathcal{D}(y_{t}, y^{ref}_{t})\right] \leq d
    \end{aligned}
\end{equation}

\finalrev{\textit{Remark}: the inclusion of the system state $x$ and the true fault $z$ is needed since $s_t$ ought to be a sufficient statistic for the future evolution of the environment. However, this information is not available to the agent. Therefore, during both training and deployment phases, we operate with a masked partial state $s'_t$ excluding these elements. Fig. \ref{fig:framework} illustrates both training and deployment setups. }

\section{Actuator Faults in Linear Systems}
\label{sec:framework application}

\subsection{Passive Component Derivation}
\label{subsec:pd derivation}

For systems modelled by Eq. \ref{eq:system equations linear}, assuming $A$, $B$ and $u_t$ to be known, a passive observer can be derived, which provides estimates for $x_t$ and $z_t$ at time $\tau$ of the form:
\begin{equation}
\begin{aligned}
    x^{est}_{t,\tau}\sim \mathcal{N}\left(\mu_{x_{t}}^{(\tau)}, \Sigma_{x_{t}}^{(\tau)}\right) \\
    z^{est}_{t,\tau}\sim \mathcal{N}\left(\mu_{z_{t}}^{(\tau)}, \Sigma_{z_{t}}^{(\tau)} \right).
\end{aligned}
 \end{equation}
Update rules for distribution parameters are derived in closed form using Bayesian inference. If omitted, assume $\tau = t$.\\
Consider $B^*_{t} = B \operatorname{diag}(u_{t})$ and Eq. \ref{eq:system equations linear} expressed as:
\begin{equation}
\label{eq:system simplified}
    x_{t+1} = A x_{t} + B^*_{t}z_{t} + w_t .
\end{equation}
Direct substitution of the estimates at time $t$ in Eq. \ref{eq:system simplified} yields: 
\begin{equation}
    \begin{aligned}
        \mu_{x_{t+1}}^{(t)} &= A \, \mu_{x_{t}}^{(t)} + B^*_{t} \, \mu_{z_{{t}}}^{(t)}\\
        \Sigma_{x_{t+1}}^{(t)} &= A \, \Sigma_{x_{t}}^{(t)} \, A^T + B^*_{t} \, \Sigma_{z_{t}}^{(t)} \, {B^*_{t}}^T + \Sigma_{w}.
    \end{aligned}
\end{equation}
\noindent To account for $y_{t+1}$, the density term $p(x_{t+1} \mid y_{t+1})$ can be directly evaluated applying Bayes' theorem, resulting in:
\begin{equation}
    \begin{aligned}
        \Sigma_{x_{t+1}}^{(t+1)} &= \left((\Sigma_{x_{t+1}}^{(t)})^{-1} + C^T \Sigma^{-1}_{v} C \right)^{-1} \\
        \mu_{x_{t+1}}^{(t+1)} &= \Sigma_{x_{t+1}}^{t+1}\left((\Sigma^{(t)}_{x_{t+1}})^{-1} \mu_{x_{t+1}}^{(t)} + C^T \Sigma^{-1}_{v} \,{y}_{t+1} \right). 
    \end{aligned}
\end{equation}
Applying Woodbury matrix identity \cite{woodbury1950inverting} and performing some algebraic manipulations, a more numerically stable set of equations can be obtained:
\begin{equation}
\label{eq:state estimate update rules}
    \begin{aligned}
        K_x(t) &= \Sigma_{x_{t+1}}^{(t)} C^T \left( \Sigma_{v} + C \Sigma_{x_{t+1}}^{(t)} C^T\right)^{-1}\\
        \Sigma_{x_{t+1}} &= \left(I -K_x(t) \; C \right) \Sigma_{x_{t+1}}^{(t)} \\
        \mu_{x_{t+1}} 
        &= \mu_{x_{t+1}}^{(t)} + K_x(t) \left({y}_{t+1} - C \mu_{x_{t+1}}^{(t)}\right).\\
    \end{aligned}
\end{equation}
\noindent The procedure up to this point is equivalent to the derivation of a standard Kalman filter, where the uncertainty in the fault vector $z_t$ adds to the existing process noise.

Parameter update rules for the fault estimate can also be derived from Eq. \ref{eq:system simplified}.
Substituting best estimates for all terms that do not depend on $z_t$ yields:
\begin{equation}
    B^*_t \, z_t \sim N\left(\mu_{x_{t+1}}^{(t+1)} - A \,\mu_{x_{t}}^{(t)}  \, , \Sigma_{x_{t+1}}^{(t+1)}  + A \Sigma_{x_t}^{(t)}  A^T + \Sigma_{w} \right)
\end{equation}
This represents a way to measure $z_t$, analogously to how the term $C x_t \sim  N({y}_{t} \, , \Sigma_{w})$ represents a measurement for $x_t$. 
Therefore, using the previous results in Eq. \ref{eq:state estimate update rules} and substituting the correct mean and variances: 
\begin{equation}
    \begin{aligned}
        K_z(t) &= \Sigma_{z_{t}}^{(t)} {B^*_t}^T \left( \Sigma_{x_{t+1}}^{(t+1)} + \Sigma_{x_{t+1}}^{(t)} \right)^{-1} \\
        \Sigma_{z_{t}}^{(t+1)} &= (I -K_z(t) B^*_t ) \Sigma_{z_{t}}^{(t)}\\
        \mu_{z_{t}}^{(t+1)} &= \mu_{z_{t}}^{(t)} + K_z(t) (\mu_{x_{t+1}}^{(t+1)} -  \mu_{x_{t+1}}^{(t)}).\\   
    \end{aligned}
\end{equation}
Note that this improves the estimate for the parameter at time $t$, using information obtained from $y_{t+1}$ at the following time step.
To obtain $z_{t+1}^{(t+1)}$, some assumptions on the fault evolution need to be made. In the case of $\xi_t\sim \mathcal{N}\left(\mu_{\xi_t} , \Sigma_{\xi_t} \right)$ the parameter updates become:
\begin{equation}
        \Sigma_{z_{t+1}}^{(t+1)} = \Sigma_{z_{t}}^{(t+1)} + \Sigma_{\xi_{t}}, \;
        \mu_{z_{t+1}}^{(t+1)} = \mu_{z_{t}}^{(t+1)} + \mu_{\xi_{t}}.
\end{equation}
Notably, when $\mu_{\xi_t}=0$, the update reduces to adding a bias term to the estimate covariance. This simple addition produces remarkably stable filters that can perform reasonably even if the fault dynamics do not match the assumptions (i.e. jump processes) as shown in \ref{sec:case study}.

\subsection{\ac{FIERL} application}
\label{subsec:framework application}

Since the considered passive component provides estimates in the form of Gaussian distributions, the selected reward metric is the negative expected squared norm of the fault estimation error, which admits the closed form:
\begin{equation*}
\begin{aligned}
    \mathcal{R}(s_t, a_t, s_{t+1}) &= - \mathbb{E}_{z\sim z^{est}_{t}}\left(\lVert z^{true}_t - z\rVert^2_2 \right) \\
    & = - \operatorname{trace} (\Sigma_{z_t}) -\rVert z_t^{true}-\mu_{z_t} \rVert^2_2.
\end{aligned}
\end{equation*}
\revised{The cost function counts the number of times the adherence to the reference exceeds a desired tolerance $\Delta y^{max}$:
\begin{equation}
\label{eq:A2_costfunction1}
    C(s_t, a_t, s_{t+1}) = \begin{cases}
    0 & \text{if } \lVert y_{t} - y^{ref}_{t} \rVert_\infty \leq \Delta y^{max} \\
    1 & \text{if } \lVert y_{t} - y^{ref}_{t} \rVert_\infty > \Delta y^{max}
\end{cases}.
\end{equation}}
The full state vector incorporates the state and fault estimate distribution parameters, the system output and reference:
\begin{equation*}
    s'_t = 
    \begin{bmatrix}
        {\mu}_{x_t} & \operatorname{triu}({\Sigma}_{x_t}) & {\mu}_{z_t} &\operatorname{triu}({\Sigma}_{z_t}) & y^{ref}_t & y_t
    \end{bmatrix}.
\end{equation*}
\noindent Therefore the optimal policy becomes:
\begin{equation}
    \centering
    \begin{aligned}
    \pi_{\theta^*} = &\arg\max_{\pi_\theta} \mathbb{E}_{\tau\sim\pi_\theta}\left[\sum_{t=0}^{\infty} -\gamma^t \; \mathbb{E}_{z\sim z^{est}_{t}}\left(\lVert z^{true}_t - z\rVert^2_2 \right) \right]. \\
    \text{s.t.} \quad &  \mathbb{E}_{\tau\sim\pi_\theta}\left[\sum_{t=0}^{\infty} \gamma_c^t \; \mathbf{1}_{(\Delta y^{max}, \infty)}\left(\lVert y_{t} - y^{ref}_{t} \rVert_\infty\right)\right] \leq d
    \end{aligned}
\end{equation}
\revised{In this case the parameter $d$ represents the allowed tracking bound violation rate. For $d=0$ it must be enforced at all times.}

\section{Case Study}
\label{sec:case study}

\begin{figure}
\centering
\includegraphics[width=\columnwidth]{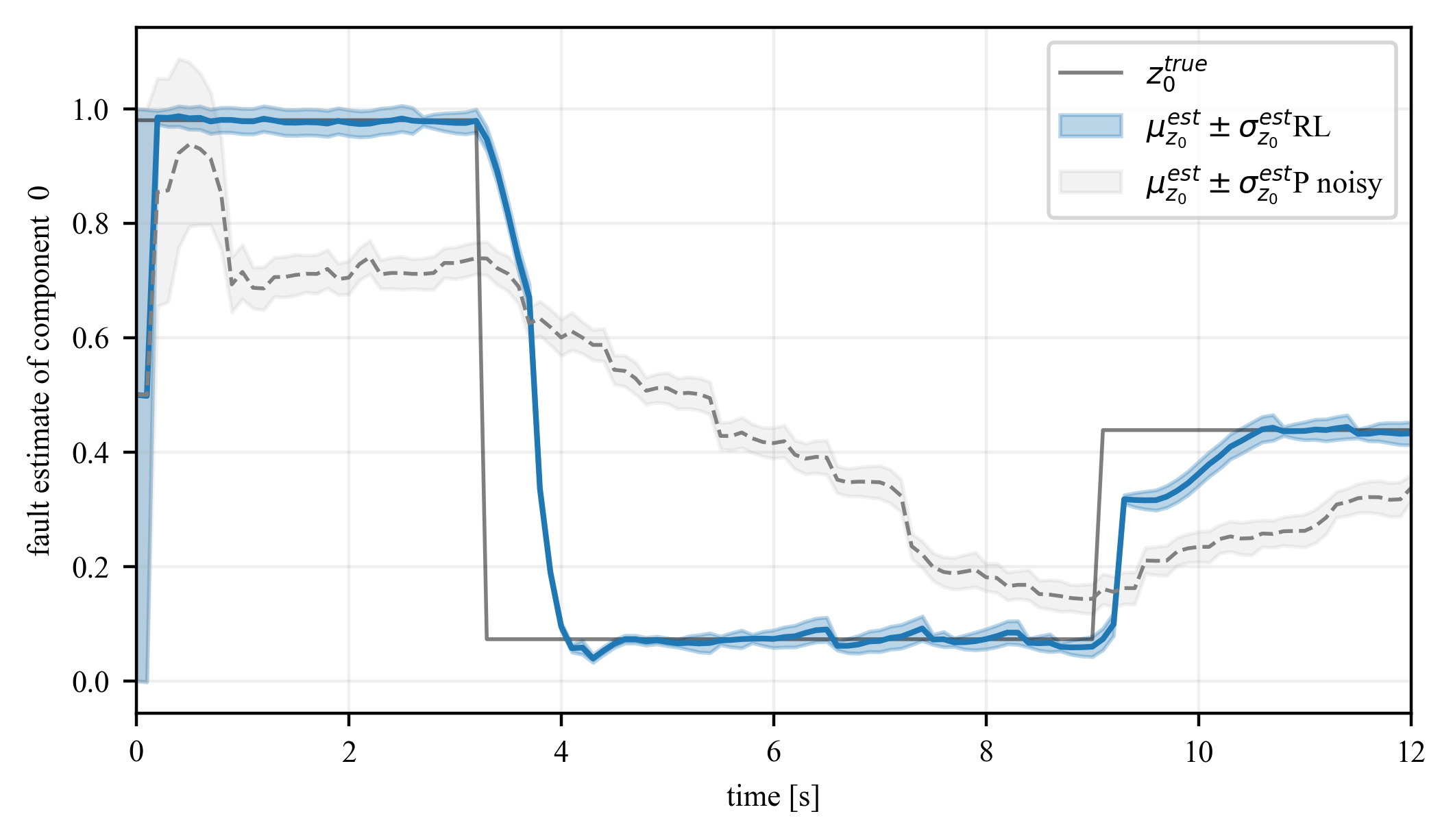}
\includegraphics[width=\columnwidth]{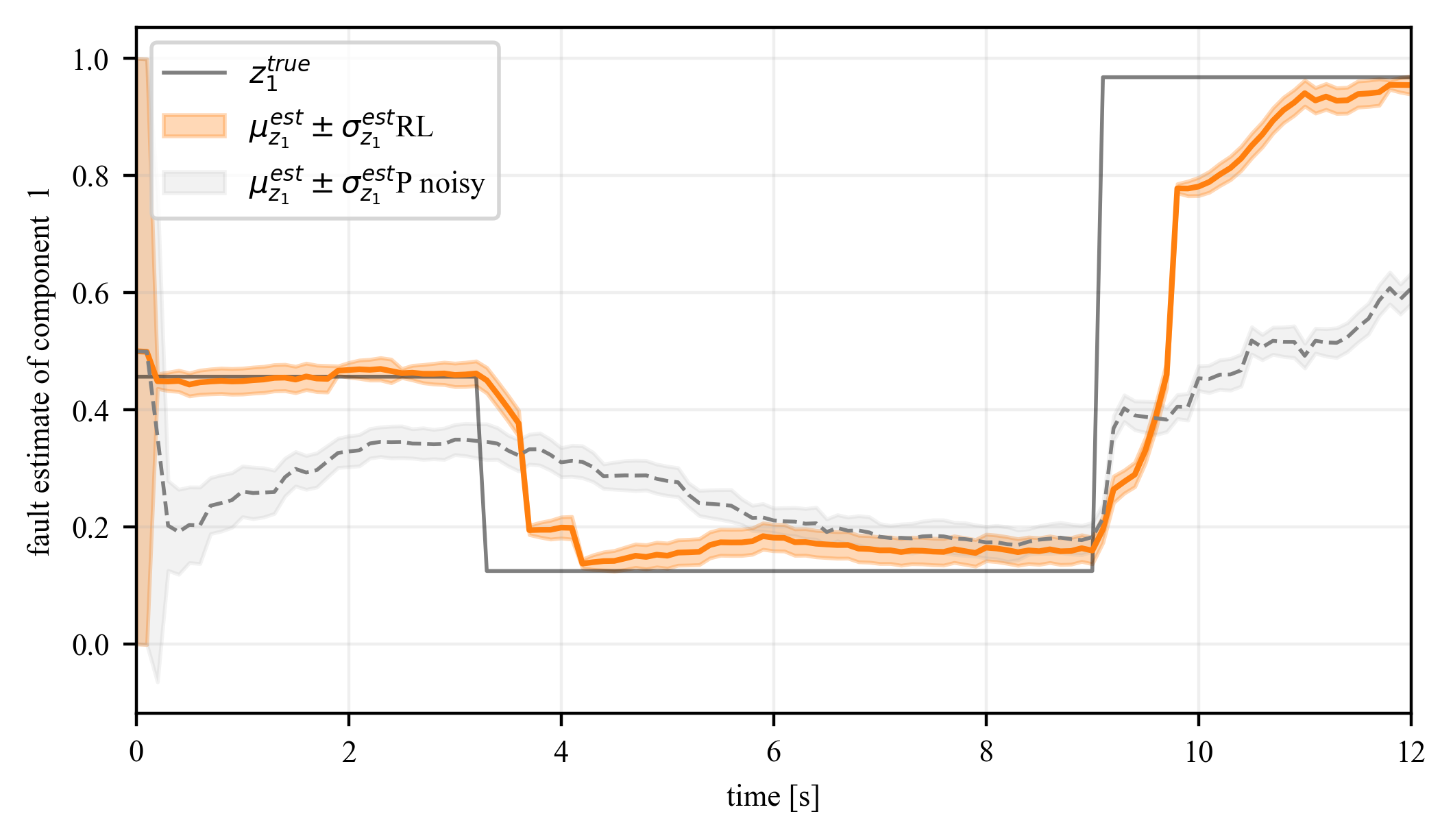}

\includegraphics[width=\columnwidth]{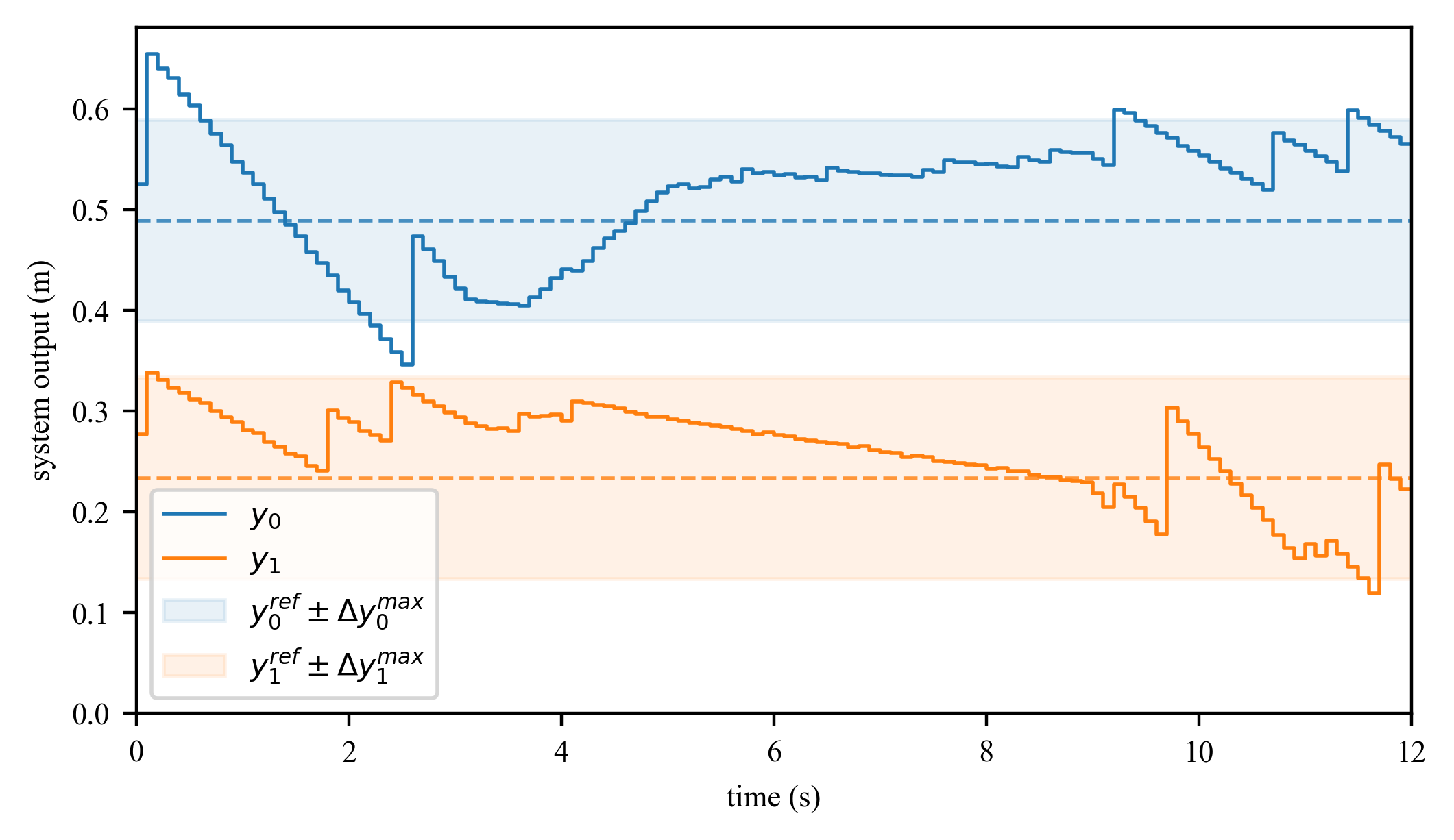}
\includegraphics[width=\columnwidth]{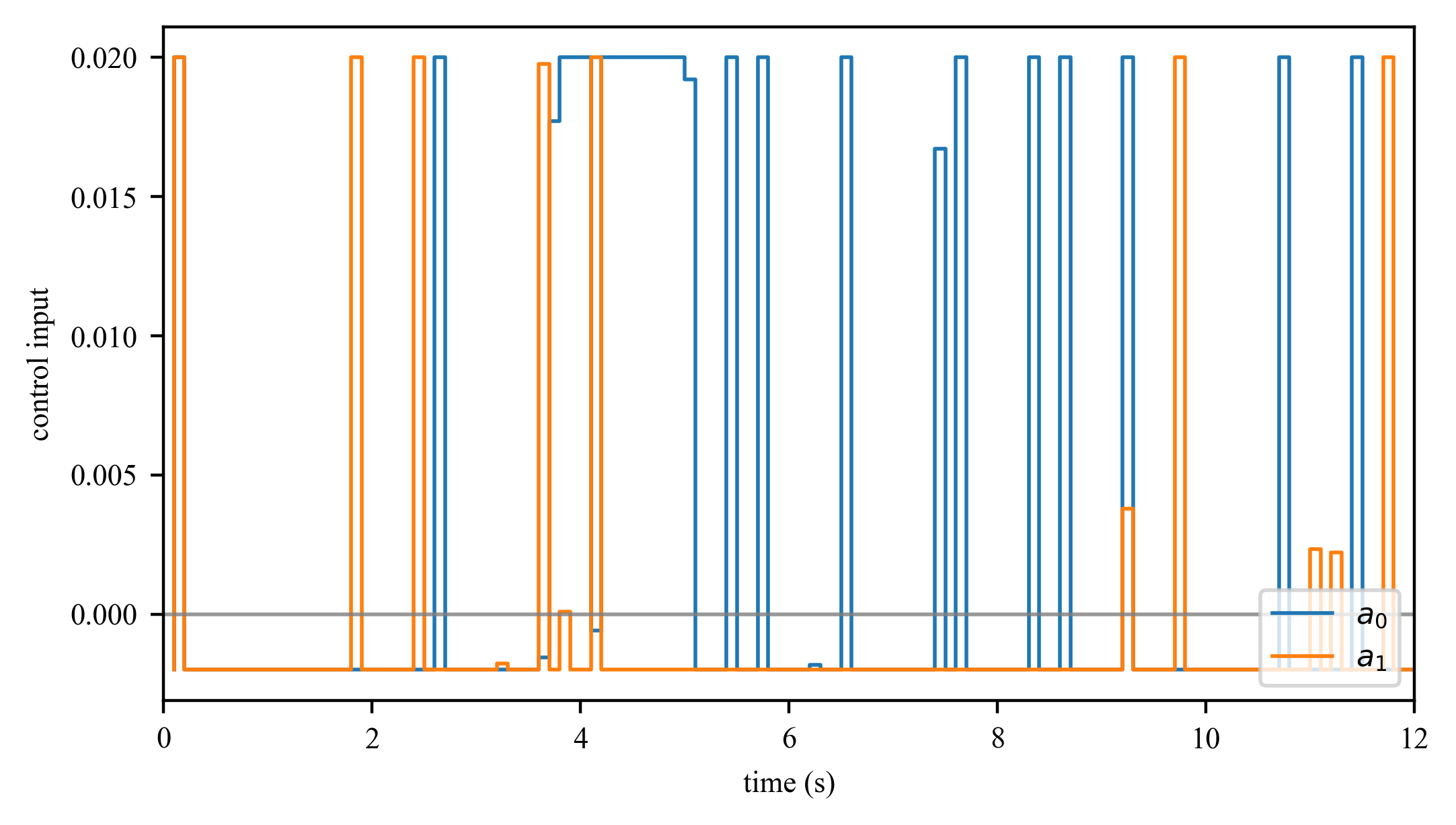}

\caption{
\revised{Experimental results for a random testing episode. The top two panels show the evolution of fault estimate distributions for \ac{FIERL} compared to the naive approach. The fault estimate obtained with the RL policy promptly approaches the true fault values after jumps, and generally it tends to be more accurate, while also maintaining a tighter distribution.\\
Because of the passive component design, single bursty actions provide more information on the fault mode than applying the equivalent control over a longer period of time. This is in accordance with the \ac{RL} policy behaviour, depicted in the fourth panel, where a large action is followed by a sequence of small negative ones in order to meet the tracking performance requirements (shown in panel 3).}
}
\label{fig:experiments}
\end{figure}

\begin{figure}
\centering
\includegraphics[width=\columnwidth]{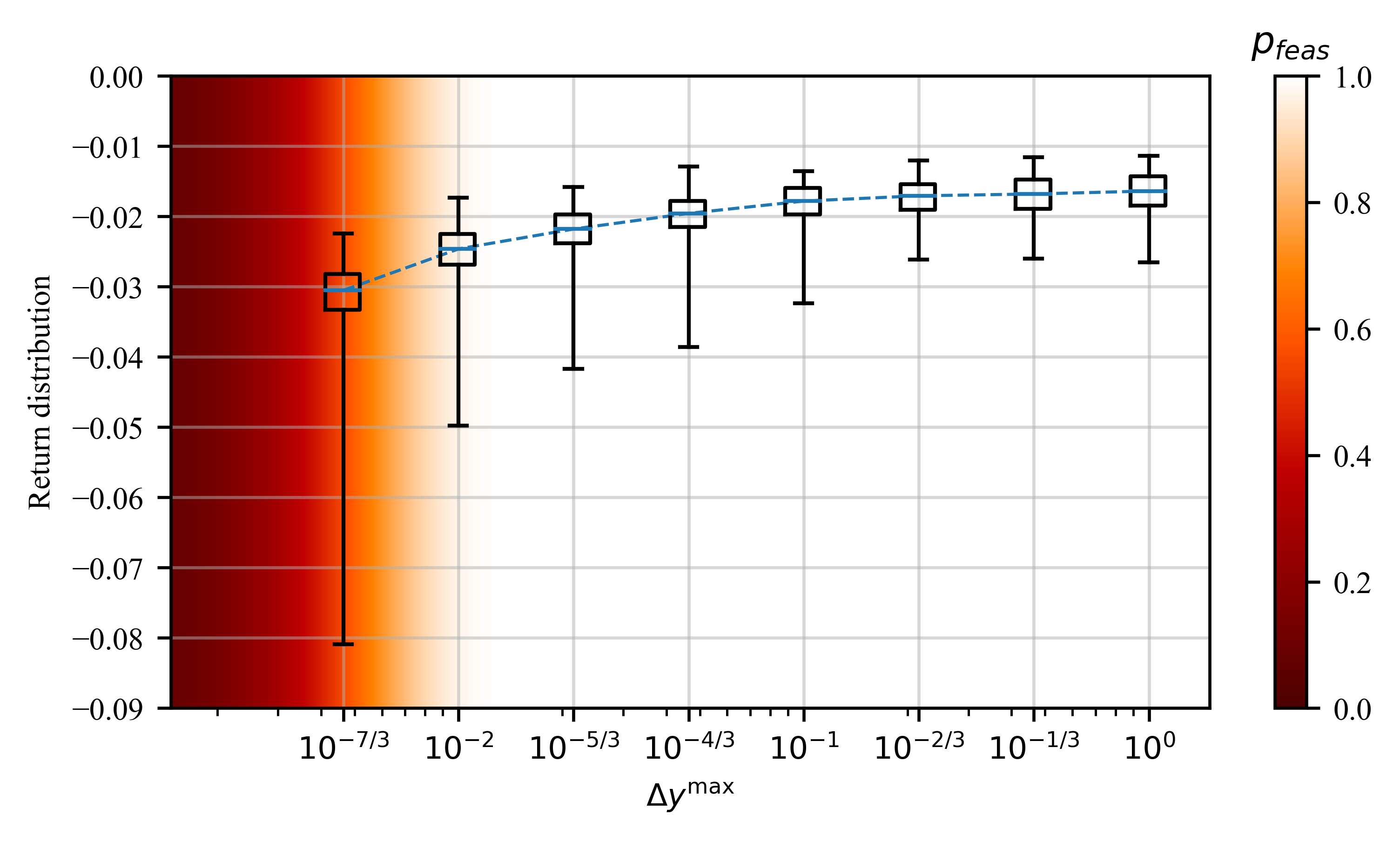}
\caption{\revised{Diagnosis performance as a function of tracking threshold $\Delta y^{max}$. 
Each boxplot represents the average return distribution in test episodes for \ac{FIERL} with the corresponding control requirements. The color-map shows the probability of violating the constraints purely due drifts caused by process noise. As $\Delta y^{max}$ shrinks, the diagnosis performance quickly degrades, and the training becomes unstable. This is because the starting policy, no matter how conservatively it is initialized, is almost certainly unfeasible, which is needed to ensure convergence bounds \cite{achiam2017constrained}}.}
\label{fig:beta-tradeoff}
\end{figure}

In this section \ac{FIERL} is applied to the three-tank benchmark system.
The system model, in the form of Eq. \ref{eq:system equations linear} is derived in the Appendix of \cite{en13174475}. 
The system state represents the liquid levels in the tanks, and system matrices are obtained through linearization around the equilibrium point ${x^*=[0.489, 0.2332, 0.3611]^T}$ meters 
discretized using a sampling time of $T_S=0.1$ seconds. 
\revised{The covariance for i.i.d process and measurement noise are, respectively, ${\Sigma_w = 10^{-8}}$ and ${\Sigma_v =10^{-6}}$}. 
The control input ${u_t \in \mathbb{R}^2}$ is limited to ${U=[-0.002, 0.02]}$.
The system output is the first and the second components of the state. This choice is driven by the absence of direct control over the third component.

\finalrev{In order to assess its effectiveness, \ac{FIERL} policy is compared to the naive approach of a propoprtional controller where the control input is subject to random perturbation}
\revised{${\Delta u^P_t \sim \mathcal{U}([-K_p, K_p])}$. The gain $K_p$ is tuned in order to achieve the best diagnosis performance while respecting the same tracking constraints of the \ac{RL} policy.} This simple approach is surprisingly effective, guaranteeing reasonable control performance thanks to the proportional controller, while acquiring significantly more information on the fault state.
In fact, if the domain of the input were symmetric (as it was originally presented in \cite{en13174475}), the naive approach would be a remarkably close approximation of the optimal policy, making it difficult to appreciate the \finalrev{FIERL} approach potential. \vale{This is the primary reason for the modification.}

The \ac{RL} agent is trained for 1000 policy updates, each one performed using data collected across 90 short episodes of length $N_D=40$ steps (4 seconds).
%The threshold on the constraint is $d=6$, i.e. $d=0.15N_D$.
At each episode the true fault vector, which remains constant for the entire duration, is sampled uniformly in ${[0,1]\times[0,1]}$, while the true initial liquid level is sampled uniformly in the ball of radius $\Delta y^{max}=0.1$ centered in $x^*$. 
The cost constraints is set to ${d = 6 = 0.15N_D}$. The passive component is initialized with a wide uninformative prior ${z^{est}_0\sim \mathcal{N}((0.5,0.5),\operatorname{diag}(1,1))}$ 
and the bias term of the estimated covariance is $\Sigma_{\omega_t}=10^{-3}$.

The testing setup consists of long episodes with a variable number of steps ranging from 90 to 180 (instead of 40 used in training). The true fault trajectory is assumed to change abruptly, re-sampling the fault vector at random time steps, guaranteeing that each fault persists for at least 30 steps, allowing the module enough time to produce a reliable estimate. 
For the \ac{RL} policy, averaging over 10000 test episodes yields an expected return per step of \revised{${R_{RL} = (-1.458 \pm 0.562) \cdot 10^{-2}}$, and an expected C-return per step of ${C_{RL} = (1.178 \pm 1.145)\cdot 10^{-1}}$}. For the proportional controller the expected returns are significantly lower at \revised{${R_{P} = (-3.892 \pm 1.477) \cdot 10^{-2}}$, ${C_{P} = (0.030 \pm 2.18)\cdot 10^{-3}}$}. 

\revised{Finally Fig. \ref{fig:experiments} illustrates a single test episode in detail, while Fig. \ref{fig:beta-tradeoff} describes the trade-off between tracking constraints and diagnosis performance.}

\finalrev{Code of the experiments is available at: \\ \url{https://github.com/davidesartor/FIERL}}.

\section{CONCLUSIONS}
\label{sec:conclusions}
This work introduces a general application of \ac{CRL} to enhance the performance of passive diagnosis modules. The integration of \ac{FIERL} with the application-specific \ac{PFD} component derived in section \ref{subsec:pd derivation} highlights several important features of the method. Most importantly, it does not require perfect knowledge of the system state, relying on state estimates instead. Additionally, unlike the majority of existing literature, this strategy can effectively handle a continuous spectrum of faults or, more in general, any stochastic fault dynamics that can be effectively simulated. Furthermore, \ac{FIERL} can, in principle, be directly applied to non-linear time varying systems. This might result in more complicated \ac{PFD} components, but would not affect the \ac{RL} loop.
Finally, \ac{FIERL} is robust to unseen fault dynamics, as shown in section \ref{sec:case study} where, even by training on stationary faults only, experimental results prove its effectiveness in handling transient faults. This is particularly important when considering the possibility of continuous monitoring. 

Conversely, \ac{RL} is computationally intensive, demanding accurate simulations (and accurate system models) to ensure the correct dynamics are learned. Moreover, determining when convergence has been achieved throughout the entire environment state domain is not trivial nor is it possible to ensure full convergence is always reached. \revised{Theoretical guarantees can be obtained by assuming perfect knowledge of the value function, which is unrealistic for practical applications \cite{Neu2017AUV}}.
Lastly, a noteworthy observation pertains to the cost function. The one used in \ref{sec:case study} aims to limit the average number of constraint violations to a maximum of $d$ times. However, this is only enforced on average across multiple episodes. 
This might result in a objective miss-specification, since the optimal policy might significantly over-violate the constraint in certain domain regions, offset by others where the violation rate is almost zero. 

Directions for further investigation include the replacement of the indicator function with a cost that discourages multiple constraint violation in the same episode and testing on more challenging scenarios such as non-linear or over-actuated system, and to more generic fault dynamics.

% \bibliographystyle{IEEEtran}
% \bibliography{references1}{}

\end{document}